\documentclass[letterpaper, 10 pt, conference]{ieeeconf}  

\IEEEoverridecommandlockouts                              

\usepackage{graphicx}
\usepackage{booktabs}
\usepackage{amsmath}

\usepackage{amssymb}

\usepackage{subcaption}
\usepackage{graphicx}
\usepackage{pgf}            
\usepackage[export]{adjustbox} 
\usepackage{graphicx}

\usepackage{algorithm}
\usepackage{algorithmic}

\usepackage{subcaption}

\usepackage{hyperref}  
\usepackage{cite}
\usepackage{subcaption}
\usepackage{multirow}
\usepackage{makecell}

\usepackage{amsthm}
\newtheorem{insight}{Empirical Insight}
\renewcommand{\rm}{r_{\textnormal{max}}}

\newcommand{\bR}{\mathbb{R}}

\newcommand{\cA}{\mathcal{A}}

\newcommand{\cF}{\mathcal{F}}

\newcommand{\cM}{\mathcal{M}}
\newcommand{\cP}{\mathcal{P}}

\newcommand{\cS}{\mathcal{S}}
\newcommand{\cU}{\mathcal{U}}

\newcommand{\cB}{\mathcal{B}}

\usepackage[usenames,dvipsnames,svgnames,x11names]{xcolor}

\title{\LARGE \bf
Reliable Policy Iteration:  Performance Robustness Across Architecture and Environment Perturbations}

\author{S.R. Eshwar$^{1}$ Aniruddha Mukherjee$^{2}$ Kintan Saha$^{1}$ Krishna Agarwal$^{1}$ \\  Gugan Thoppe$^{1, \dagger}$  Aditya Gopalan$^{1}$   Gal Dalal$^{3}$
\thanks{$^{1}$Indian Institute of Science, Bangalore, India.}
\thanks{\hspace{-1.5em}\tt\small (eshwarsr, kintansaha, krishnaagarw, gthoppe, aditya)@iisc.ac.in}
\thanks{$^{2}$Kalinga Institute of Industrial Technology, Bhubaneswar, India. {\tt\small mukh.aniruddha@gmail.com}}%
\thanks{$^{3}$NVIDIA Research, Israel. {\tt\small gdalal@nvidia.com}}%
\thanks{$^\dagger$GT's research is supported in part by grants from the Walmart Centre for Tech Excellence at IISc, the Indo-French Centre for the Promotion of Advanced Research Grant (CEFIPRA; Project 7102-1), Kotak IISc FinTech Grant, DST-SERB's Core Research Grant (CRG/2021/008330), and the Pratiksha Trust Young Investigator Award.}%
}

\begin{document}

\maketitle
\thispagestyle{empty}
\pagestyle{empty}

\begin{abstract}
In a recent work, we proposed Reliable Policy Iteration (RPI), that restores policy iteration's monotonicity-of-value-estimates property to the function approximation setting. Here, we assess the robustness of RPI's empirical performance on two classical control tasks---CartPole and Inverted Pendulum---under changes to neural network and environmental parameters. Relative to DQN, Double DQN, DDPG, TD3, and PPO, RPI reaches near-optimal performance early and sustains this policy as training proceeds. Because deep RL methods are often hampered by sample inefficiency, training instability, and hyperparameter sensitivity, our results highlight RPI's promise as a more reliable alternative.
\end{abstract}


\section{Introduction}
Policy Iteration (PI) \cite{howard1960dynamic} is an iterative method for solving sequential decision-making problems. It requires (i) full knowledge of the model---i.e., the state transition dynamics and the instantaneous rewards under different actions, and (ii) access to the exact value function of each policy, i.e., a quantitative evaluation of each strategy's long-term performance. In this setting, every successive policy that PI identifies is provably better in terms of its value function.

While PI guides the design of several Reinforcement Learning (RL) methods, such as DQN, Double DQN \cite{dqn}, DDPG \cite{ddpg}, TD3 \cite{td3}, and PPO \cite{schulman2017proximal}, real-world applications bring forth two key challenges. First, the environment model is unknown and the state-action dynamics and rewards can only be sampled via interactions. Second, the state and action spaces are very large and sometimes even infinite, necessitating the use of (possibly high-capacity) function approximators for representing value functions. In such challenging settings, PI's theoretical guarantees no longer apply since the value estimates need not faithfully reflect the current policy's true performance. In fact, these poor estimates often misguide policy updates, leading to suboptimal or divergent behavior. This theory-practice gap has been the source of several problems in RL, including sample inefficiency, training instability, and hyper-parameter sensitivity.

Motivated by the above gap, we recently proposed Reliable Policy Iteration (RPI) \cite{reliablecritics}---shown in Algorithm~\ref{alg:RPI} here. RPI retains PI's greedy policy improvement while reformulates value estimation as a novel Bellman-inequality-constrained optimization problem, in lieu of plain projection or Bellman-error minimization. With a known model, RPI's estimates are monotonically non-decreasing and provide a pointwise lower bound on the current policy's value function \cite[Theorem~3.1]{reliablecritics}. In this way, RPI restores PI's monotonicity guarantee \textit{under arbitrary function approximation}.

Here, we evaluate the performance robustness of a model-free variant of RPI---also proposed in \cite{reliablecritics}---on CartPole-v1 and InvertedPendulum-v5. These tasks serve as representative benchmarks for discrete and continuous control, respectively. Our study comprises two experiments. First, we test RPI's performance stability against the aforementioned RL methods by systematically varying the capacity of the function approximator. Second, to assess generalizability, we examine whether the performance trends observed above persist even when the environments' physical parameters (such as pole mass, cart mass, and gravity) are perturbed. RPI's theoretical formulation explicitly models the impact of function approximation \cite{reliablecritics}. Hence, we expect it to exhibit greater performance stability than previous RL approaches.

Our key observations are as follows. 
\begin{enumerate} 

\item \textbf{Resilience to Function Approximator Architecture:} 
Our findings reveal that, across a range of function approximator choices, RPI consistently learns a near-optimal policy using a small number of samples and more or less holds this performance as more samples are subsequently obtained. In contrast, other methods either are slower to learn or show higher training instability in one or both the environments, especially with low-capacity neural networks.

\item \textbf{Robustness to Environment Modifications:} When the environment's core physical parameters such as pole mass, cart mass, and gravity are changed, we find that RPI maintains its aforementioned learning trend relative to other methods. We emphasize that all algorithms here utilize the hyperparameters that were obtained by tuning on unmodified environments.

\end{enumerate}

The remainder of this paper is organized as follows. Section \ref{sec2:rpi} briefly reviews the RPI framework. Section \ref{sec:experiments} presents our comprehensive empirical evaluation. Finally, Section \ref{sec:conclusion} concludes with a summary of our findings.

\section{RPI Algorithm Details}
\label{sec2:rpi}

In this section, we briefly review the RPI framework, which we introduced in \cite{reliablecritics}. We first describe RPI's model-based version and then present its model-free variant. For details on RPI's theoretical analyses and the derivation of the model-free loss \eqref{eq:rpi_loss}, please see the original work \cite{reliablecritics}.

\subsection{Setup}
We consider a stationary Markov Decision Process (MDP) denoted by $\cM = (\cS, \cA, \cP, r, \gamma).$ Here, $\cS$ and $\cA$ represent finite sets of states and actions, respectively, and the cardinalities $|\cS| = S$ and $|\cA| = A.$ Further, $\cP$ denotes the transition dynamics with $\cP(s' \mid s, a)$ indicating the probability of transitioning to state $s'$ upon taking action $a$ in state $s$. Finally, $r: \cS \times \cA \to \bR$ assigns a scalar reward to each state-action pair, and $\gamma \in [0, 1)$ denotes the discount factor.

Let $\Delta(\cU)$ denote the set of probability distributions over a finite set $\cU$. A stationary policy $\mu: \cS \to \Delta(\cA)$ maps each state to a distribution over actions. The action-value function (or Q-function) $Q_\mu: \cS \times \cA \to \bR$ of a policy $\mu$ is defined as $Q_\mu(s, a) := \mathbb{E} \left[ \sum_{t = 0}^\infty \gamma^t r(s_t, a_t) ,\middle|, s_0 = s, a_0 = a \right]$, where  $s_{t+1} \sim \cP(\cdot \mid s_t, a_t)$ and $a_{t+1} \sim \mu(\cdot \mid s_{t+1})$ for $t \geq 0.$ Similarly, the Bellman operator $T_\mu: \bR^{SA} \to \bR^{SA}$ corresponding to a fixed policy $\mu$ is defined as
\begin{multline}
\label{eq:T_mu}
(T_\mu Q)(s, a) \\ :=  
r(s, a) + \gamma \sum_{s', a'} \cP(s' \mid s, a)\mu(a' \mid s') Q(s', a').
\end{multline}

\begin{figure*}[t]
    \centering
    \includegraphics[width=\linewidth]{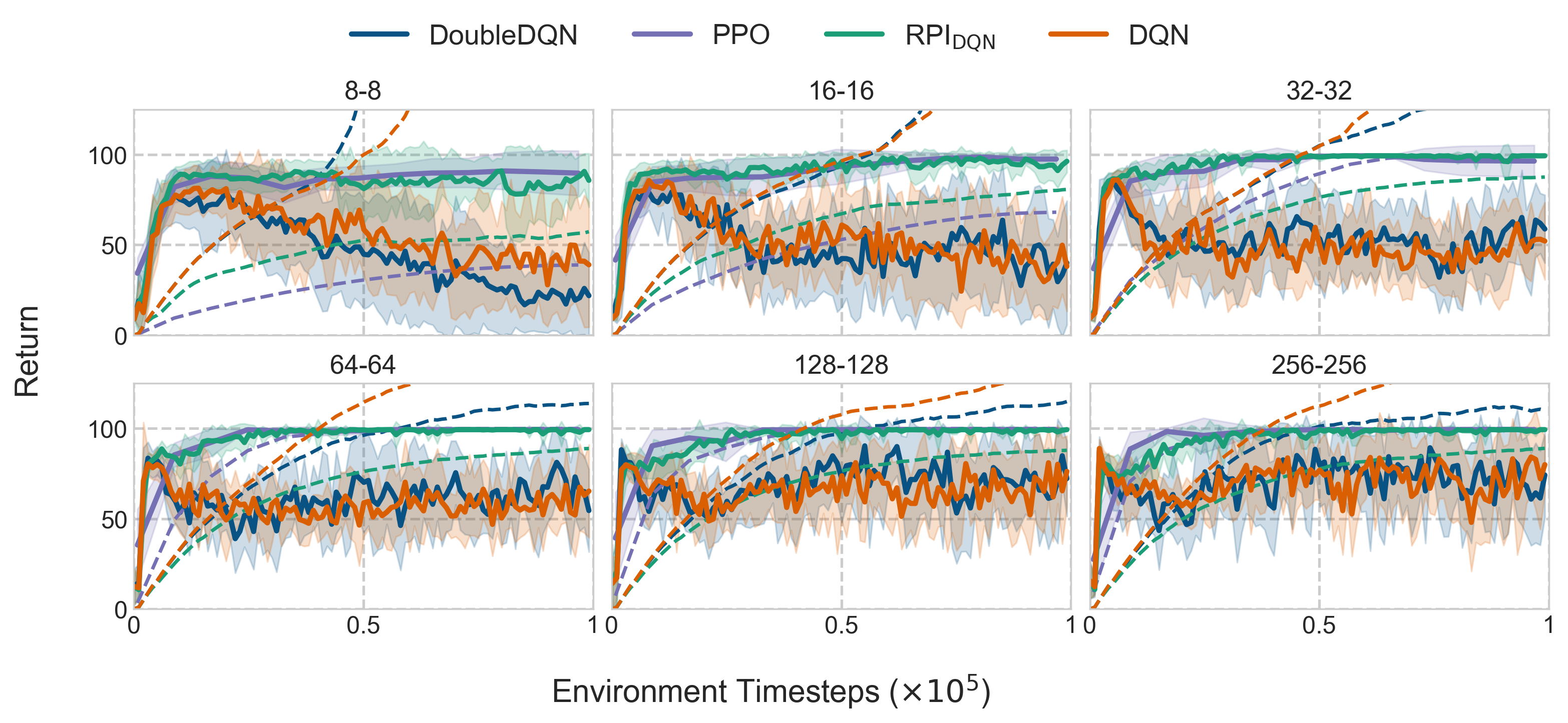}
    \caption{Training curves for CartPole-v1 across varying network capacities (sub-plot titles indicate neurons per layer). Solid lines: true returns; dashed lines: critic estimates; shaded regions: mean $\pm$ std over 10 runs. \textbf{Summary:} RPI\textsubscript{DQN} achieves performance comparable to PPO while its critic estimates remain a lower bound to true return. DQN and Double DQN consistently overestimate values and the performance degrades under low-capacity networks. Also, note that DQN and Double-DQN's performances often drop after initially identifying a reasonably good policy.
    }
    \label{fig:cartpole_learning_curves}
\end{figure*}

\begin{figure*}[t]
    \centering
    \includegraphics[width=\linewidth]{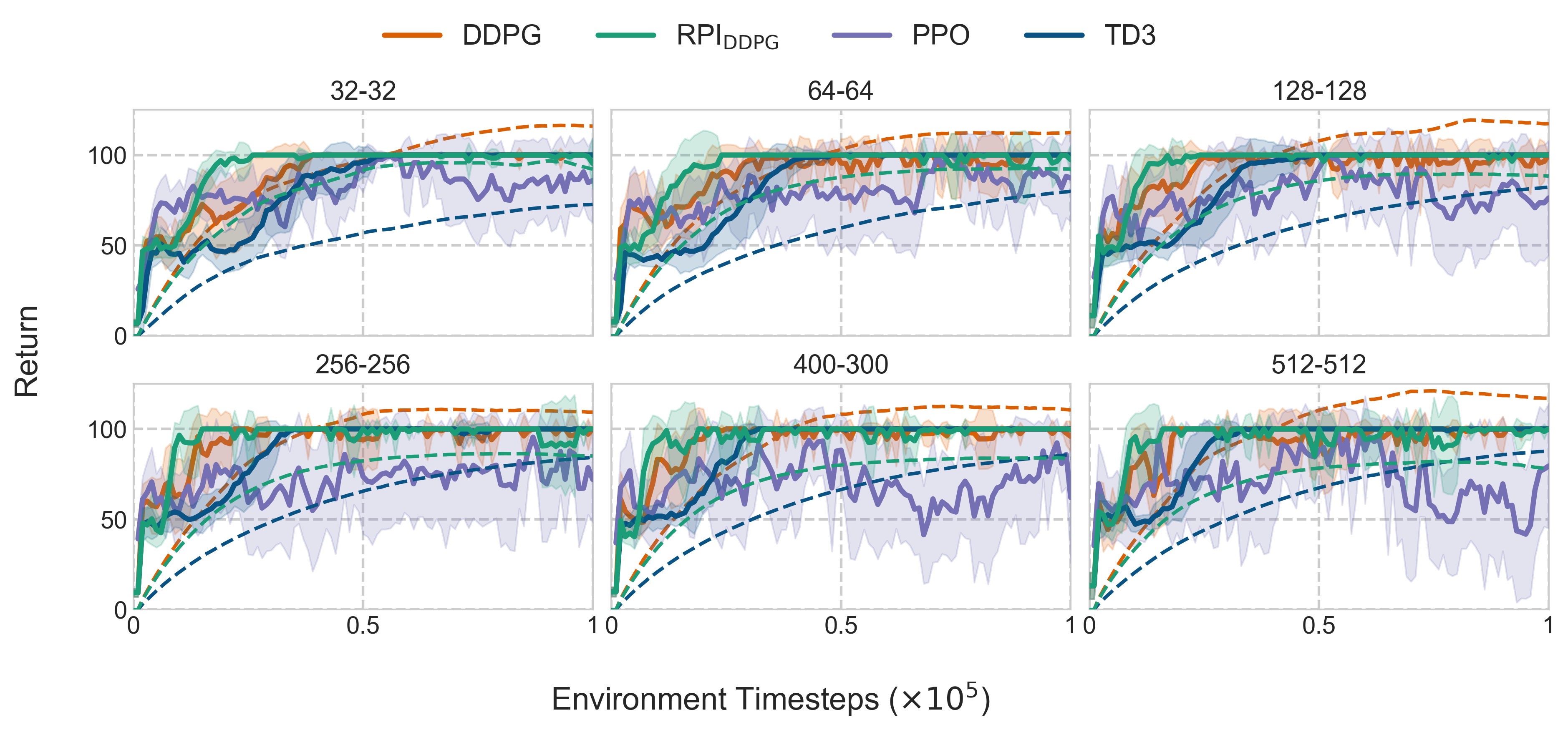}
    \caption{Training curves for InvertedPendulum-v5 across varying network capacities (sub-plot titles indicate neurons per layer). Lines/shading follow  the convention in Fig.~\ref{fig:cartpole_learning_curves}. \textbf{Summary:} RPI\textsubscript{DDPG} learns fastest, achieving high AUC while its critic estimates remain a lower bound. DDPG's critic overestimes, while TD3 learns slowest but converges stably. PPO struggles to compete and exhibits high variance.}
    \label{fig:inverted_pend_learning_curves}
\end{figure*}

\subsection{Model-Based RPI Algorithm}

RPI's pseudocode is given in Algorithm~\ref{alg:RPI}. Given a function approximation class $\mathcal{F},$ an initial policy $\mu_0$ and its corresponding Q-function estimate $f_0 \in \mathcal{F}$, RPI alternates between policy evaluation and policy update. While these two steps are common to any PI-style method, the core novelty is in the policy evaluation step. Unlike standard approaches that minimize Bellman or projection errors, RPI formulates policy evaluation as a one-sided Bellman-inequality-based constrained optimization problem. 

Specifically, in the update step---as is done in classical PI \cite{howard1960dynamic}---the current policy $\mu_k$ is replaced with the one that is greedy with respect to the new value estimate $f_{k + 1}.$ On the other hand, during evaluation, RPI introduces  \eqref{e:GUIDE.PE.Opt} so as to find an estimate $f$ that respects  $T_{\mu_k} f \geq f$ and is, coordinate-wise, the maximal vector dominating $f_k.$ Since $T_\mu$ is monotone and a contraction for any $\mu$ \cite{bertsekas1996neuro}, the constraint $T_{\mu} f \geq f$ implies $Q_{\mu} \geq f$ (though not conversely). Consequently, \eqref{e:GUIDE.PE.Opt}  is designed to compute the tightest in-class lower bound to $Q_{\mu_k}$ that also preserves $f \geq f_k.$ 

\begin{algorithm}[t!]
   \caption{Reliable Policy Iteration (RPI)}
   \label{alg:RPI}
\begin{algorithmic}
   \STATE {\bfseries Input:} FA class $\cF,$ policy $\mu_0,$ an initial approximation $f_0 \in \cF$ of $Q_{\mu_0},$ and a norm $\|\cdot\|$ 
   \FOR{$k = 0, 1, 2$ ... until convergence}
   \STATE \textbf{Policy Evaluation}:
    \begin{equation}
    \label{e:GUIDE.PE.Opt}
        \begin{aligned}
            f_{k+1} \in  \underset{f \in \cF}{\arg\max} \quad & \|f - f_k\| \\
            \text{s.t.} \quad & T_{\mu_k} f  \geq f \geq f_k. 
        \end{aligned}
    \end{equation}
    \STATE \textbf{Policy Improvement}: 
    \begin{equation}
    \begin{split}
        \mu_{k + 1} \in & \  \{\mu: \mu \text{ is a deterministic policy} \\
        & \quad \text{that is greedy w.r.t. } f_{k + 1} \}
    \end{split}
    \end{equation}
    \ENDFOR
\end{algorithmic}
\end{algorithm}

As shown in \cite[Theorem 3.1]{reliablecritics}, RPI's key advantage is that its sequence of value estimates $(f_k)$ is non-decreasing, and that each $f_k \geq Q_{\mu_k}$ (pointwise). Together, these properties extend---for the first time---the monotonicity guarantees of tabular PI to the function approximation setting.

\subsection{Model-Free RPI Critic Loss}
The model-based formulation in Algorithm~\ref{alg:RPI} is impractical for many real-world applications, as it requires knowledge of the transition dynamics (to compute the Bellman operator $T_{\mu_k}$) and involves solving a constrained optimization over the entire state-action space. To overcome these limitations, a practical, model-free critic loss was derived in \cite{reliablecritics}. The key idea is to rewrite the constrained optimization problem as an unconstrained one: fold the constraints into the objective function using penalty functions and use sample-based approximations for the latter.

The resulting critic loss function is
\begin{multline}
\label{eq:rpi_loss}
    \mathcal{L}^{\text{RPI}}(f) := \frac{1}{|\cB|}\sum_{i=1}^{|\cB|} \Bigg[ -c\cdot f(s_i, a_i) + \lambda_1 \Big[ f(s_i, a_i) - y_i \Big]_+ \\
    \quad + \lambda_2 \Big[q_\text{min} - f(s_i, a_i) \Big]_+ \Bigg].
\end{multline}
Here, $\mathcal{B}$ is a minibatch of transitions, $f$ is the critic network output, $[x]_ = \max(x, 0)$ is the ReLU function, and $c,\lambda_1, \lambda_2 > 0$ are hyperparameters, of which $\lambda_1$ and $\lambda_2$ penalize constraint violations. The value $y_i,$ which is sample estimate of $T_\mu f (s_i, a_i)$ (with $\mu$ being greedy with respect to (a delayed copy of) $f$), is computed analogously to standard DQN or DDPG, i.e., by using a target network to ensure stability. The term $q_{\text{min}}$ is a fixed, predefined lower bound on the Q-values, ensuring the estimates improve relative to this fixed baseline. 

The loss in \eqref{eq:rpi_loss} trains the critic to produce nondecreasing value estimates that lower-bound the true discounted returns. It can drop in as a direct replacement for the Mean Squared Bellman Error in deep RL algorithms such as DQN \cite{dqn} and DDPG \cite{ddpg}, yielding the RPI variants RPI\textsubscript{DQN} and RPI\textsubscript{DDPG}.

\begin{table*}[t]

\centering

\caption{Comparison of algorithms across different network architectures on CartPole and Inverted Pendulum. We report mean $\pm$ standard deviation over 10 seeds. Final Performance is the average return over the final 10\% of training steps. AUC (×10$^6$) measures sample efficiency. Steps to solve (×10$^3$) denotes the timesteps required to first solve the environment.}

\label{tab:network_arch_comparison}

\begin{tabular}{lllcccccc}

\toprule

\textbf{Environment} & \textbf{Algorithm} & \textbf{Metric} & \multicolumn{6}{c}{\textbf{Network Architecture}} \\

\cmidrule(lr){4-9}

 & & & 8-8 & 16-16 & 32-32 & 64-64 & 128-128 & 256-256 \\

\midrule

\multirow{12}{*}{\textbf{CartPole}} & \multirow{3}{*}{Double DQN} & Final Perf. & 21.6 ± 17.9 & 38.1 ± 22.8 & 57.7 ± 16.3 & 68.2 ± 10.6 & 69.3 ± 12.0 & 70.8 ± 9.2 \\

 &  & AUC  & 4.7 ± 1.5 & 5.0 ± 1.2 & 5.4 ± 0.6 & 6.1 ± 0.5 & 6.8 ± 0.5 & 7.0 ± 0.3 \\

 &  & Steps to solve  & 52.6 ± 32.9 & 41.0 ± 23.2 & 38.0 ± 22.6 & 33.7 ± 21.5 & 30.0 ± 18.1 & 17.1 ± 13.5 \\

\cmidrule(lr){2-9}

 & \multirow{3}{*}{PPO} & Final Perf. & \textbf{89.8 ± 12.3} & \textbf{97.5 ± 2.8} & 96.4 ± 8.8 & \textbf{99.3 ± 0.0} & \textbf{99.3 ± 0.0} & \textbf{99.3 ± 0.0}\\

 &  & AUC  & \textbf{8.5 ± 0.4} & 9.0 ± 0.3 & 9.3 ± 0.2 & \textbf{9.5 ± 0.1} & \textbf{9.5 ± 0.1} & \textbf{ 9.6 ± 0.1} \\

 &  & Steps to solve  & \textbf{41.3 ± 25.0} & 41.0 ± 21.7 & \textbf{19.0 ± 8.9} & 16.9 ± 5.9 & \textbf{14.8 ± 9.4} & 16.2 ± 8.7 \\

\cmidrule(lr){2-9}

 & \multirow{3}{*}{DQN} & Final Perf. & 42.4 ± 27.1 & 39.2 ± 20.5 & 52.3 ± 10.3 & 61.3 ± 11.7 & 69.6 ± 9.1 & 75.0 ± 10.8 \\

 &  & AUC  & 5.6 ± 1.6 & 5.3 ± 0.8 & 5.0 ± 0.8 & 5.7 ± 0.6 & 6.5 ± 0.4 & 7.0 ± 0.4 \\

 &  & Steps to solve  & 55.5 ± 17.7 & \textbf{28.5 ± 20.4} & 29.7 ± 23.2 & 41.5 ± 29.0 & 39.4 ± 25.6 & 27.5 ± 13.7 \\

 \cmidrule(lr){2-9}

 & \multirow{3}{*}{RPI\textsubscript{DQN}} & Final Perf. & 85.4 ± 13.7 & 94.6 ± 4.7 & \textbf{99.0 ± 0.7} & 99.0 ± 0.5 & 98.7 ± 0.5 & 98.9 ± 0.3 \\

 &  & AUC  & 8.3 ± 0.9 & \textbf{9.1 ± 0.4} & \textbf{9.5 ± 0.1} &\textbf{ 9.5 ± 0.0} & 9.4 ± 0.1 & 9.3 ± 0.1 \\

 &  & Steps to solve  & 66.1 ± 24.1 & 35.9 ± 13.2 & 20.4 ± 4.1 & \textbf{16.4 ± 3.1} & 16.7 ± 4.4 & \textbf{14.7 ± 3.1} \\

\midrule

& & & \multicolumn{6}{c}{\textbf{Network Architecture}}\\

\cmidrule(lr){4-9}

 & & & 32-32 & 64-64 & 128-128 & 256-256 & 400-300 & 512-512 \\

\cmidrule(lr){2-9}

\multirow{15}{*}{\textbf{\makecell{Inverted\\Pendulum}}} & \multirow{3}{*}{DDPG} & Final Perf. & 99.8 ± 0.6 & \textbf{100.0 ± 0.0} & 97.8 ± 4.6 & 98.7 ± 2.3 & 99.1 ± 1.8 & 99.8 ± 0.5 \\

 &  & AUC  & 8.8 ± 0.3 & 9.0 ± 0.4 & 9.1 ± 0.3 & 9.3 ± 0.2 & 9.2 ± 0.2 & 9.1 ± 0.2 \\

 &  & Steps to solve  & 25.7 ± 9.3 & 18.8 ± 13.2 & 14.9 ± 6.1 & \textbf{10.0 ± 3.2} & 12.8 ± 3.3 & 11.0 ± 4.6 \\

\cmidrule(lr){2-9}

 & \multirow{3}{*}{PPO} & Final Perf. & 85.5 ± 18.4 & 88.4 ± 9.8 & 77.5 ± 21.5 & 77.3 ± 19.2 & 79.0 ± 11.3 & 59.0 ± 25.1 \\
 &  & AUC  & 8.0 ± 0.8 & 7.9 ± 0.6 & 7.7 ± 0.6 & 7.1 ± 0.7 & 6.9 ± 0.6 & 7.0 ± 0.8 \\
 &  & Steps to solve  & 42.6 ± 11.2 & 28.9 ± 10.5 & 24.2 ± 9.1 & 24.6 ± 9.0 & 22.2 ± 6.5 & 18.8 ± 4.2 \\

\cmidrule(lr){2-9}

 & \multirow{3}{*}{TD3} & Final Perf. & \textbf{100.0 ± 0.0} & \textbf{100.0 ± 0.0} & \textbf{100.0 ± 0.0} & \textbf{100.0 ± 0.1} & \textbf{100.0 ± 0.0} & \textbf{100.0 ± 0.0} \\

 &  & AUC  & 8.1 ± 0.3 & 8.3 ± 0.3 & 8.5 ± 0.3 & 8.7 ± 0.2 & 8.8 ± 0.2 & 8.9 ± 0.2 \\

 &  & Steps to solve  & 42.3 ± 9.1 & 35.9 ± 7.4 & 34.0 ± 6.0 & 30.9 ± 4.6 & 27.2 ± 4.6 & 26.2 ± 5.2 \\

 \cmidrule(lr){2-9}

 & \multirow{3}{*}{RPI\textsubscript{DDPG}} & Final Perf. & 99.1 ± 1.8 & 98.3 ± 4.1 & 99.6 ± 1.0 & 94.6 ± 12.9 & \textbf{100.0 ± 0.1} & 98.9 ± 1.8 \\

 &  & AUC  & \textbf{9.2 ± 0.1} & \textbf{9.2 ± 0.2} & \textbf{9.4 ± 0.1} & \textbf{9.4 ± 0.2} & \textbf{9.4 ± 0.1} & \textbf{9.3 ± 0.2} \\

 &  & Steps to solve  & \textbf{18.8 ± 3.4} & \textbf{15.9 ± 3.6} & \textbf{11.0 ± 2.2} & 10.5 ± 2.2 & \textbf{8.7 ± 2.1} & \textbf{8.8 ± 1.2} \\

\bottomrule

\end{tabular}

\end{table*}

\section{Experiments}
\label{sec:experiments}

In the previous section, we presented RPI's theoretical formulation, which explicitly accounts for function approximation errors. In this section, we examine whether this modeling translates into the expected empirical benefits. Specifically, we address two central questions:
\begin{enumerate}
    \item Is RPI resilient to variations in the complexity of the function approximator (neural network architecture) (Section \ref{subsec:func-approx})?
    \item Does RPI maintain its relative performance advantages when the environment's core parameters change (Section \ref{subsec:env-mod})?
\end{enumerate}

To answer the above question, we benchmark RPI in two standard control environments. For the discrete CartPole task, we compare RPI\textsubscript{DQN} to DQN \cite{dqn}, Double DQN (DDQN) \cite{van2016deep}, and PPO \cite{schulman2017proximal}. For the continuous Inverted Pendulum task, we evaluate RPI\textsubscript{DDPG} against PPO \cite{schulman2017proximal}, DDPG \cite{ddpg}, and TD3 \cite{td3}.

We evaluate these algorithms using the following metrics, which capture  learning efficiency and stability.

\noindent\textbf{Evaluation Metrics:}
(i) \textit{Final Performance}-the average return over the final 10\% of training steps, indicating asymptotic policy quality; 
(ii) \textit{Steps to Solve}-the number of timesteps required for the agent to first solve the task; and 
(iii) \textit{Area Under the Learning Curve (AUC)}-a measure of overall sample efficiency and training stability. 

We conclude the section with implementation details (Section \ref{subsec:impln-dets}).

\begin{figure*}[t]
    \centering
    \includegraphics[width=\linewidth]{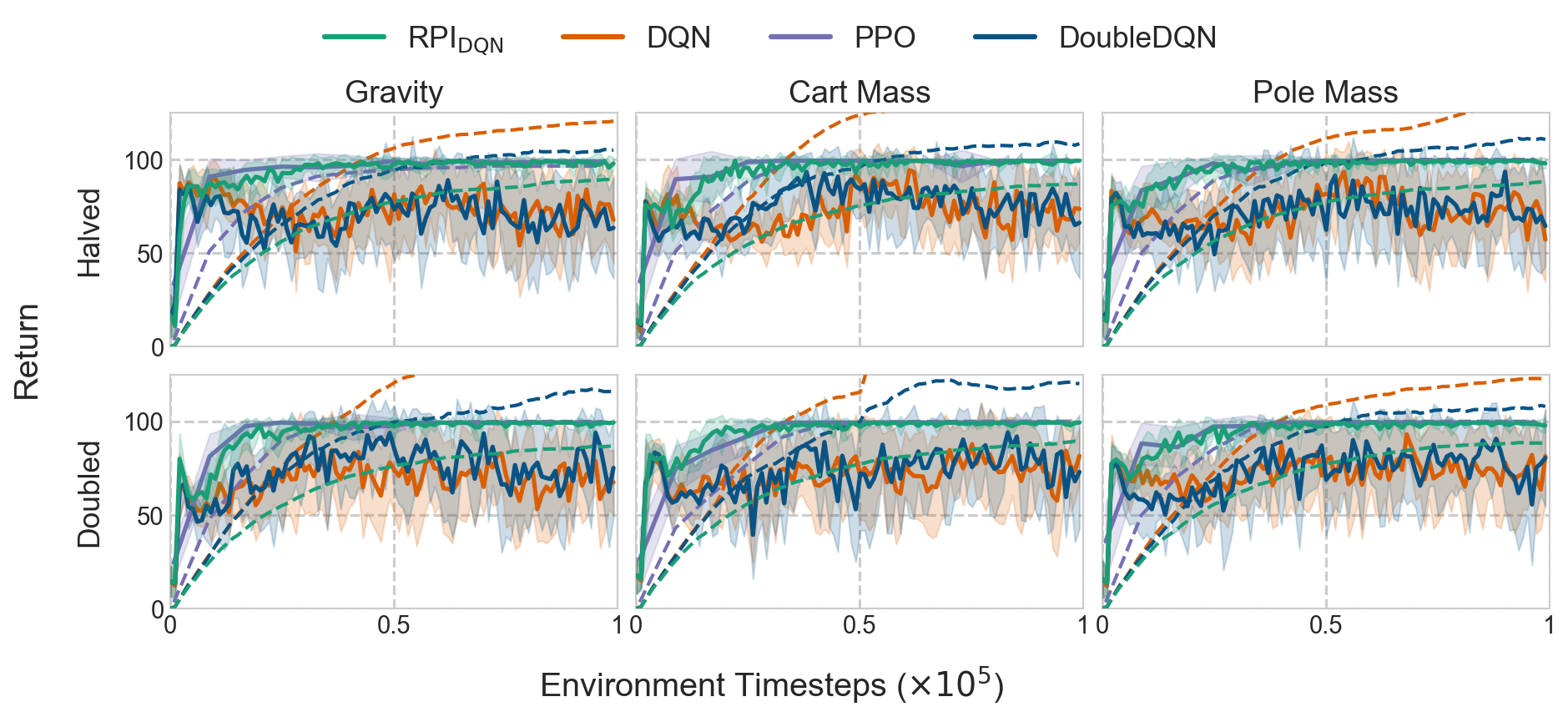}
    \caption{
    Training performance on six CartPole-v1 variants with modified physical properties. Columns: gravity, cart mass, pole mass changes; Rows: halved (top), doubled (bottom) parameters. Solid lines: discounted Monte Carlo returns; dashed lines: estimated value function; shaded areas: mean $\pm$ standard deviation over 10 runs. \textbf{Summary:}
    All algorithms generally show relatively stable performance across these environment modifications. Within this context, RPI\textsubscript{DQN}  surpasses DQN/Double DQN and matches PPO's performance. RPI\textsubscript{DQN} consistently maintains its critic estimate as a lower-bound to the true return.
    }
    \label{fig:cartpole_learning_curves_env_config}
\end{figure*}

\begin{figure*}[t]
    \centering
    \includegraphics[width=\linewidth]{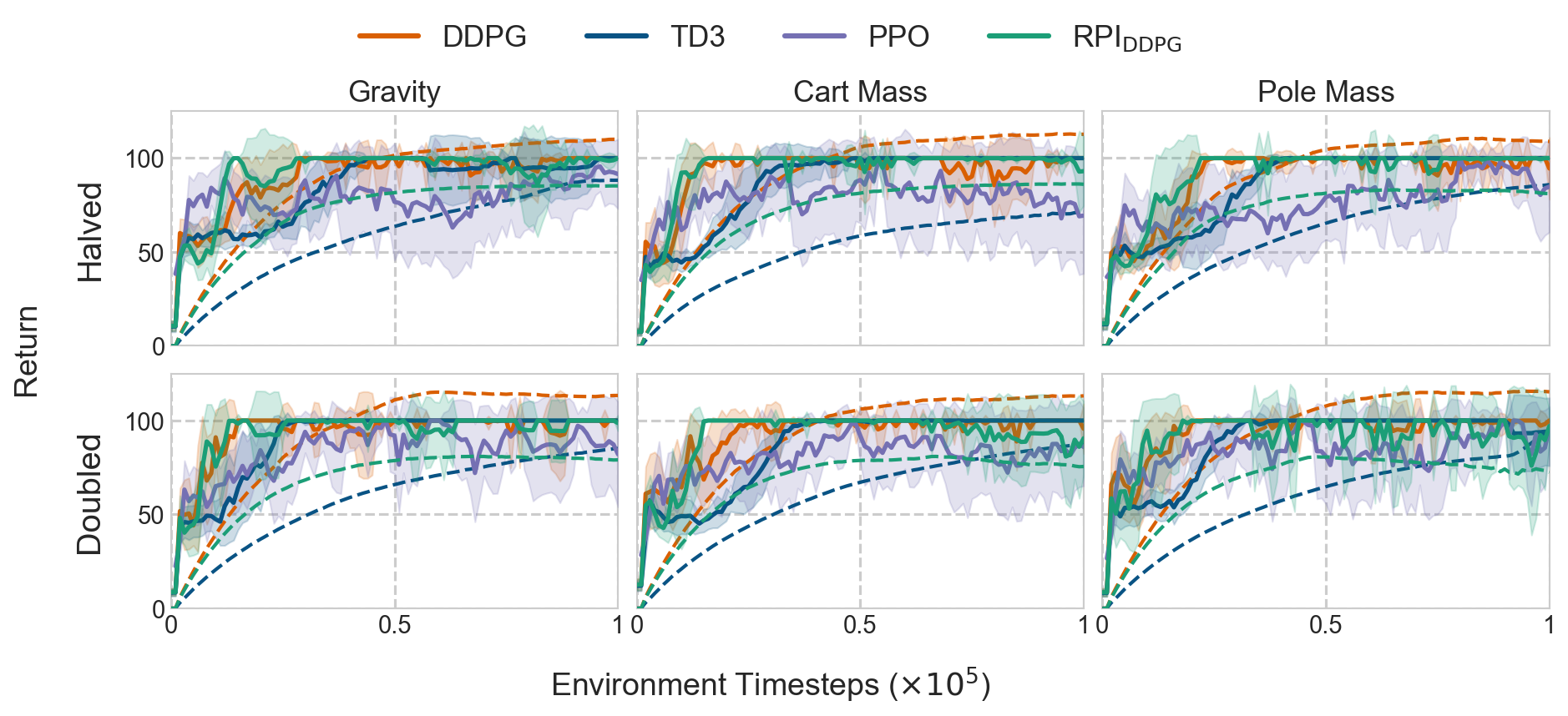}
    \caption{
    Training performance on six InvertedPendulum-v5 variants with modified physical properties. Columns: gravity, cart mass, pole mass changes; Rows: halved (top), doubled (bottom) parameters. Lines/shading follow previous convention. \textbf{Summary:} All algorithms generally show relatively stable performance across these environment modifications. Within this context, RPI\textsubscript{DDPG} learns faster than TD3/PPO and often matches DDPG, while consistently maintaining its critic as a lower-bound to the true return. TD3 learns slowly but is stable. PPO exhibits high variance.
    }
    \label{fig:inverted_pend_learning_curves_env_config}
\end{figure*}

\subsection{Resilience to function approximation}
\label{subsec:func-approx}
We now address the first question: whether RPI's explicit handling of function approximation yields resilient learning across a spectrum of function approximator complexities, from simple to powerful. Since all algorithms utilize a two-layer neural network architecture for their actor and/or critic components, we systematically vary the number of neurons per layer to study how network capacity influences learning behavior. Specifically, we consider architectures 
$\{8-8,\,16-16,\,32-32,\,64-64,\,128-128,\,256-256\}$ for CartPole-v1 and 
$\{32-32,\,64-64,\,128-128,\,256-256,\,400-300,\,512-512\}$ for InvertedPendulum-v5. 
Each configuration is trained on 10 different seeds.

Figures~\ref{fig:cartpole_learning_curves} and~\ref{fig:inverted_pend_learning_curves} show the averaged learning curves. Solid lines denote discounted Monte Carlo returns (averaged over 100 roll-outs), dashed lines denote predicted Q-values (averaged over the initial state-action pairs of the same roll-outs), and shaded region indicate one standard deviation.\\

\subsubsection{Observations in Carpole-v1} We compare RPI\textsubscript{DQN} against DQN and DDQN, which also utilize greedy policy updates. The learning curves (Figure \ref{fig:cartpole_learning_curves}) clearly illustrate the known issue of policy degradation when combining function approximation with greedy policy updates \cite{young2020understanding, gopalan2025does}. For instance, with the $8 - 8$ architecture, both DQN and DDQN initially learn, but subsequently suffer catastrophic performance degradation, dropping more than 50\% from their peak values. Similar, though less severe, degradations occur with $16-16$ and $32-32$ architectures as well. In contrast, RPI\textsubscript{DQN} demonstrates stable learning across all capacities.

\begin{insight} RPI\textsubscript{DQN} avoids the performance collapse and successfully solves the task, even in low-capacity settings. This suggests RPI effectively mitigates the instabilities associated with function approximation errors in greedy value-based methods.\end{insight}

Furthermore, the critic's behavior differs significantly between the algorithms. DQN and DDQN critics tend to overestimate the policy value (dashed lines), often exceeding the theoretical maximum return of $\approx99.34$ (corresponding to a full 500-step episode with $\gamma=0.99$) even when the policy value is suboptimal (solid line). This overestimation persists across architectures, making the critic estimates \emph{unreliable} indicators of true performance. Conversely, RPI\textsubscript{DQN}'s critic \emph{consistently} estimates a lower bound to true value across all network architectures.

\begin{insight}
The lower-bound property (value estimate $\leq$ true value), theoretically guaranteed for model-based RPI \cite{reliablecritics}, holds true in this model-free deep RL setting. As seen in the Figure \ref{fig:cartpole_learning_curves} the lower bound becomes tighter as the network capacity increases.
\end{insight}

The aggregate metrics in Table \ref{tab:network_arch_comparison} quantify the above mentioned trends. RPI\textsubscript{DQN} achieves significantly better final performance than value-based methods like DQN and DDQN (e.g., 2x and 4x respectively at $8-8$ ), leading to substantially better (1.3x to 1.8x)  sample efficiency (AUC) across all architectures. RPI\textsubscript{DQN} also becomes considerably faster (`Steps to solve') than DQN/DDQN from the 64-64 architecture onwards. Compared to PPO, RPI\textsubscript{DQN}'s AUC is highly competitive and generally comparable. Neither consistently dominates in convergence speed, but RPI\textsubscript{DQN} shows notably lower variance in solve times at higher capacities. Despite its simpler greedy policy updates, RPI achieves performance comparable to PPO, a more complex method that relies on a dedicated policy network and conservative updates.

\subsubsection{Observations in InvertedPendulum-v5} We compare RPI\textsubscript{DDPG} against DDPG, TD3, and PPO (Figure \ref{fig:inverted_pend_learning_curves}). RPI\textsubscript{DDPG} consistently demonstrates the fastest learning across most architectures. DDPG learns faster than TD3 but generally slower than RPI\textsubscript{DDPG}. In contrast, TD3 converges considerably slower but eventually achieves a highly stable policy. PPO exhibits high variance and struggles to match the performance of the other methods. Critically, both RPI\textsubscript{DDPG}'s and TD3's critic estimates are lower-bounds to the true values, contrasting sharply with DDPG's frequent and significant overestimation. The aggregate metrics (Table \ref{tab:network_arch_comparison}) confirm these findings. RPI\textsubscript{DDPG} solves the environment fastest (`Steps to solve') in most configurations, resulting in the highest sample efficiency (AUC) across all architectures. Although TD3 attains excellent final stability, its slower learning (requiring 2x-3x more time than RPI\textsubscript{DDPG}) leads to lower AUC. PPO consistently lags behind in all metrics.

\subsection{Robustness to Environment Modifications}
\label{subsec:env-mod}
We now address the second question: whether RPI maintains its relative performance advantages when environment dynamics change. We modify the core physical parameters of CartPole-v1 and InvertedPendulum-v5 while keeping all algorithm hyperparameters fixed to those tuned on the standard versions, testing algorithm generalization. 

For both environments, we perturb one parameter at a time---gravity, cart mass, or pole mass---by halving or doubling its default value. This results in six variants per environment. Each configuration is trained across 10 random seeds. Figures~\ref{fig:cartpole_learning_curves_env_config} and~\ref{fig:inverted_pend_learning_curves_env_config} present the learning curves. Lines/shading follow the previous convention described in Section \ref{subsec:func-approx}. 

Overall, the results across both CartPole (Figure \ref{fig:cartpole_learning_curves_env_config}) and Inverted Pendulum (Figure \ref{fig:inverted_pend_learning_curves_env_config}) demonstrate that the performance characteristics observed in the standard setting largely persist under these environment modifications. RPI variants (RPI\textsubscript{DQN}, RPI\textsubscript{DDPG}) consistently maintain their competitive advantages when faced with altered dynamics, using hyperparameters tuned for the default environment. Specifically, RPI\textsubscript{DQN} continues to demonstrate stable learning and achieve higher final performance compared to DQN/DDQN on CartPole. Similarly, RPI\textsubscript{DDPG} learns faster relative to TD3/PPO on Inverted Pendulum.
Furthermore, the RPI critics continue to uphold their lower-bound property across these variations, contrasting with the persistent overestimation seen in DQN/DDQN and DDPG. This highlights RPI's practical robustness for generalization within these task domains.

\subsection{Implementation Details}
\label{subsec:impln-dets}

For all experiments, we set $\gamma=0.99$. We used Stable-Baselines3 (SB3) implementation and fine-tuned hyperparameters for DQN and PPO\cite{sb3}. Since SB3 does not implement DDQN, we adapted code and hyperparameters from DQN. For DDPG and TD3 we use TD3's official implementation. For PPO in InvertedPendulum-v5, a mismatch exists as the parameters were tuned for $\gamma=0.999$, while our setup uses $\gamma=0.99$ in all other algorithms. We address this by training PPO using the finetuned parameters for $\gamma=0.999$ but evaluating the policy with $\gamma=0.99$ for consistent comparison. All experiments were conducted on a MacBook Pro with an M3 Max chip. The code for our experiments is available online.\footnote{\url{https://github.com/EshwarSR/RPI-ClassicControl}}

\section{Conclusion}
\label{sec:conclusion}
Our empirical evaluation shows that RPI outperforms or matches the considered baselines. Also,  its performance is  resilient to perturbations to function approximator and environment parameters. It notably avoids the policy degradation seen in DQN/DDQN, especially with smaller networks. 



\bibliography{refs}

\end{document}